\documentclass[lettersize,journal]{IEEEtran}
\usepackage{amsmath,amsfonts}
\usepackage{algorithm}
\usepackage{array}
\usepackage[caption=false,font=normalsize,labelfont=sf,textfont=sf]{subfig}
\usepackage{textcomp}
\usepackage{stfloats}
\usepackage{url}
\usepackage{verbatim}
\usepackage{graphicx}
\usepackage{cite}
\usepackage{multirow} 
\usepackage{booktabs}       % professional-quality tables
\usepackage{amsfonts}       % blackboard math symbols
\usepackage{nicefrac}       % compact symbols for 1/2, etc.
\usepackage{microtype}      % microtypography
\usepackage{xcolor}         % colors
\usepackage{url}  % author
\usepackage{graphicx}

\usepackage{colortbl}
\usepackage{makecell}
\definecolor{mygray}{gray}{.9}
\newcommand{\blue}[1]{\textcolor{black}{#1}}

\newcommand{\yellow}[1]{\textcolor{yellow}{#1}}
\newcommand{\pink}[1]{\textcolor{pink}{#1}}

\usepackage{bbding}
\usepackage{times}
\usepackage{epsfig}
\newcommand{\eg}{\textit{e}.\textit{g}.}  % author
\newcommand{\ie}{\textit{i}.\textit{e}.}  % author
\newcommand{\etal}{\textit{et al.}}  % author
\usepackage{algorithm, algpseudocode}
\usepackage{caption}
\usepackage{comment}
\usepackage{amsmath}
\usepackage{amssymb}

\usepackage{bm} % sff
\usepackage{color}
\definecolor{ggray}{RGB}{127,127,127}

\usepackage{wrapfig,lipsum,booktabs}
\usepackage{float}

\usepackage[colorlinks,
            linkcolor=red,
            anchorcolor=blue,
            citecolor=green
            ]{hyperref}

\begin{document}

\title{Counterfactual Co-occurring Learning for Bias Mitigation in Weakly-supervised Object Localization}


\author{Feifei Shao,
        Yawei Luo$^*$,
        Lei Chen,
        Ping Liu,
        Wei Yang,
        Yi Yang,
        Jun Xiao

\thanks{$^*$Yawei Luo is the corresponding author.}
\thanks{Feifei Shao, Yawei Luo, Yi Yang, Jun Xiao are with Zhejiang University, China.}
%  Email: sff@zju.edu.cn, yaweiluo@zju.edu.cn, yangyics@zju.edu.cn, junx@cs.zju.edu.cn.}
\thanks{Lei Chen is with FinVolution Group, China.}
% Email: chenlei04@xinye.com.}
\thanks{Ping Liu is with University of Nevada, Reno, America.}
%  Email: pino.pingliu@gmail.com.}
\thanks{Wei Yang is with Huazhong University of Science and Technology, China.}
%  Email: weiyangcs@hust.edu.cn.}
}



\maketitle

\begin{abstract}
    Contemporary weakly-supervised object localization (WSOL) methods have primarily focused on addressing the challenge of localizing the most discriminative region while largely overlooking the relatively less explored issue of biased activation---incorrectly spotlighting co-occurring background with the foreground feature. In this paper, we conduct a thorough causal analysis to investigate the origins of biased activation. Based on our analysis, we attribute this phenomenon to the presence of co-occurring background confounders. Building upon this profound insight, we introduce a pioneering paradigm known as Counterfactual Co-occurring Learning (CCL), meticulously engendering counterfactual representations by adeptly disentangling the foreground from the co-occurring background elements. Furthermore, we propose an innovative network architecture known as Counterfactual-CAM. This architecture seamlessly incorporates a perturbation mechanism for counterfactual representations into the vanilla CAM-based model. By training the WSOL model with these perturbed representations, we guide the model to prioritize the consistent foreground content while concurrently reducing the influence of distracting co-occurring backgrounds. To the best of our knowledge, this study represents the initial exploration of this research direction. Our extensive experiments conducted across multiple benchmarks validate the effectiveness of the proposed Counterfactual-CAM in mitigating biased activation.
\end{abstract}

\begin{IEEEkeywords}
  Co-occurring Background, Feature Disentangling, Counterfactual Representation, Weakly-supervised Object Localization.
\end{IEEEkeywords}

\section{Introduction}
\begin{figure*}[t]
    \centering
    \includegraphics[width=1.0\linewidth]{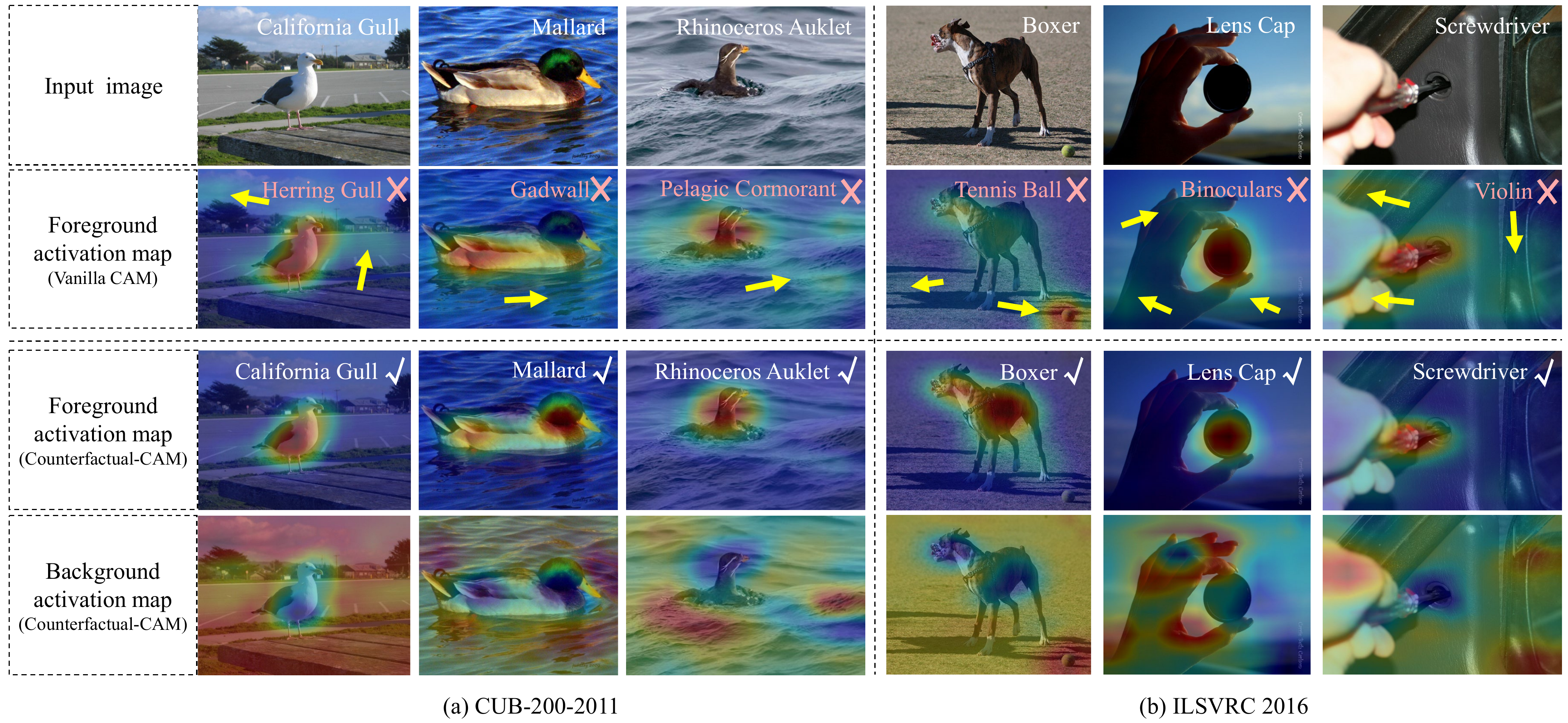}
    % \vspace{-0.5em}
    \caption{Given an input image, we visualize the foreground detected by the vanilla CAM and Counterfactual-CAM, respectively, as well as the complementary background decoupled from Counterfactual-CAM. The \pink{pink labels} and \yellow{yellow arrows} indicate the incorrect prediction category and the regions suffering from ``biased activation'', respectively.}
    \label{intro}
    \vspace{-1.0em}
  \end{figure*}

% \IEEEPARstart{A}{s} an important part of multimodal learning, cross-media reasoning has received unprecedented attention, such as Visual Question Answering (VQA)~\cite{le2021multi, hong2020selective, hong2019exploiting}, visual captioning~\cite{yang2019image, ding2020stimulus, he2019vd}, cross-media retrieval~\cite{zhang2013fusing, huang2017cross} and so on. 

Recently, the techniques based on deep convolutional neural networks (DCNNs)~\cite{simonyan2014very,he2016deep,luo2021category,luo2018macro, shao2022active,luo2024LDMC} promote the multi-modal learning, such as visual question answering (VQA)~\cite{song2023recovering, wu2023resolving, wen2023test}, scene graph generation (SGG)~\cite{li2022devil, li2023label, zhang2023end}, visual captioning~\cite{zhu2023prompt, zhao2023boosting, jing2023memory} and so on. Object detection~\cite{zhao2019object, liu2020deep, chen2021binet, fang2021ibnet} --- locating and classifying object instances in an image --- which is not only a traditional computer vision task but also a foundation technique for these multi-modal tasks connecting vision and other modalities. 

Weakly-supervised object localization (WSOL) focuses on localizing target objects in images using only image-level labels~\cite{everingham2010pascal, russakovsky2015imagenet, lin2014microsoft}. Previous approaches ~\cite{selvaraju2017grad, diba2017weakly, shao2021improving, xie2021online, kim2022bridging, wu2022background, shao2023further} have relied on class activation maps (CAMs)~\cite{zhou2016learning} to segment the highest activation area as a coarse object localization. However, these CAM-based methodologies, which are trained using image-level labels, often encounter difficulty in distinguishing between the object foreground and its co-occurring background. This issue is commonly referred to as the ``biased activation'' problem~\cite{zhang2020causal, shao2021improving}. As depicted in Figure~\ref{intro}, prior CAM-based techniques may incorrectly activate adjacent background regions, leading to erroneous classification and localization.

To address the ``biased activation'' issue, some methods~\cite{zhang2020causal, shao2021improving} have employed the structural causal model to explore the causality among the image, context, and image label. These studies have revealed the context serves as a confounder, leading the detection model to learn spurious correlations between pixels and labels. Based on the investigation, CONTA~\cite{zhang2020causal} utilizes backdoor adjustment~\cite{pearl2016causal} and the do-operator $P(Y|do(X))$ to mitigate the confounding effect of context on images. The objective is to isolate the pure causality between images and labels. Similarly, CI-CAM~\cite{shao2021improving} incorporates causal intervention into the WSOL model by using a causal context pool to address the entangled context issue. However, these approaches assume that all pertinent confounding variables have been correctly measured in the causal analysis. Neglecting unmeasured confounders can result in biased predictions and incomplete mitigation of the confounding effects~\cite{kallus2021causal, diaz2013sensitivity, zhang2018addressing}. Nevertheless, comprehensively pinpointing all the confounders remains challenging in complex scenarios. 

Inspired by existing causal intervention methods~\cite{zhang2020causal, shao2021improving}, we analyze the causality between image feature, foreground, background, and image labels. Our analysis identifies the background as a confounder that triggers the issue of ``biased activation''. In contrast to the aforementioned approaches~\cite{zhang2020causal, shao2021improving}, we propose to solve the ``biased activation'' problem stemming from the co-occurring background through counterfactual learning. Compared to causal intervention~\cite{zhang2020causal,shao2021improving}, counterfactual learning avoids the necessity to pinpoint all relevant confounding variables and offers a more explicit capability to tackle the co-occurring context. 

A counterfactual refers to a hypothetical situation that deviates from the actual course of events~\cite{pearl2018book}. By exploring counterfactuals, we can simulate scenarios in which the co-occurring background factors are automatically altered while maintaining the foreground content unchanged. Furthermore, training the model with these counterfactual scenarios accompanied by correct labels can naturally guide the model to focus on the constant foreground content while disregarding the varying background information.

Based on the counterfactual insight, we propose a novel Counterfactual Co-occurring Learning (CCL) paradigm to mitigate the negative influence of the co-occurring background in the WSOL task. More concretely, we design a Counterfactual-CAM network by introducing a counterfactual representation perturbation mechanism to the vanilla CAM. This mechanism comprises two primary steps, \ie, co-occurring feature disentangling and counterfactual representation synthesis. In the first step, a carefully designed co-occurring feature decoupler separates foreground and background features, ensuring both feature groups are orthogonal and semantically interpretable. To achieve this, we develop a new decoupled loss to control the co-occurring feature disentangling process. In the second step, the decoupled feature groups from the co-occurring feature disentangling are employed to generate counterfactual representations. These representations pair constant foreground features with various backgrounds, effectively breaking the co-occurring relationship between foreground and background. 

\begin{figure*}[t]
  \centering
  \includegraphics[width=1.0\linewidth]{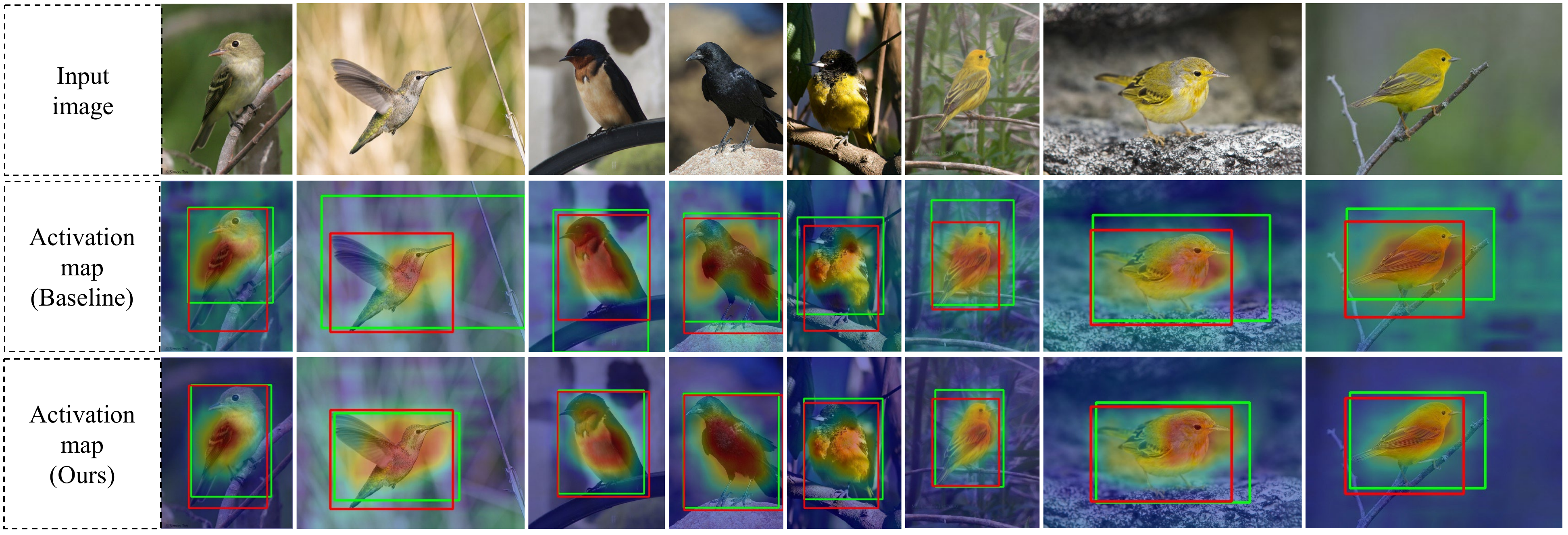}
  % \vspace{-0.5em}
  \caption{Qualitative object localization results compared with the baseline method on the CUB-200-2011 dataset. The predicted bounding boxes are in green, and the ground-truth boxes are in red.}
  \label{loc_map_cub}
  % \vspace{-0.5em}
\end{figure*}

\begin{figure*}[t]
  \centering
  \includegraphics[width=1.0\linewidth]{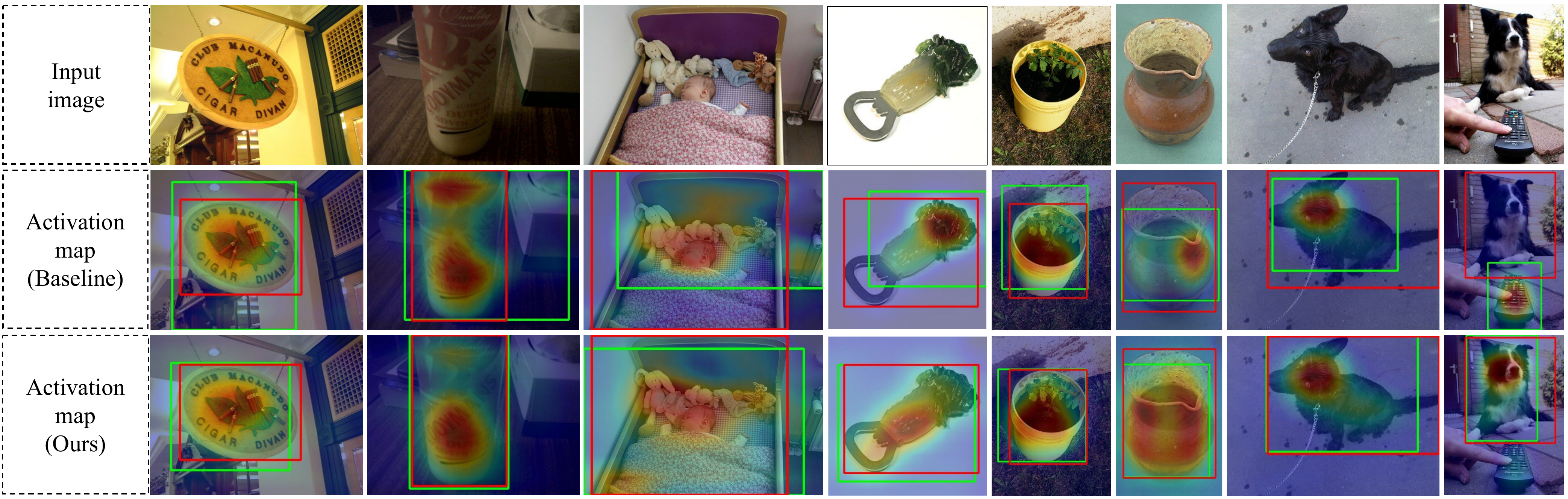}
  % \vspace{-0.5em}
  \caption{Qualitative object localization results compared with the baseline method on the ILSVRC 2016 dataset. The predicted bounding boxes are in green, and the ground-truth boxes are in red.}
  \label{loc_map_imagenet}
  \vspace{-1.0em}
\end{figure*}

Training the detection model with these synthesized counterfactual representations compels the model to prioritize constant foreground content while disregarding multifarious background information. As illustrated in Figure~\ref{intro}, Figure~\ref{loc_map_cub} and Figure~\ref{loc_map_imagenet}, this approach exhibits exceptional performance in contrast to vanilla CAM and baseline approaches, surpassing them by selectively highlighting the foreground area without affecting the background. Consequently, our method effectively addresses the ``biased activation'' problem. 

In summary, the contributions of this paper are as follows.
\begin{itemize}
  \item We propose a novel Counterfactual Co-occurring Learning (CCL) paradigm, in which we simulate counterfactual scenarios by pairing the constant foreground with unrealized backgrounds. This approach pairs constant foreground with unrealized backgrounds. As far as we know, it represents the first attempt to use counterfactual learning for mitigating co-occurring background effects in WSOL.
  \item We design a new network, dubbed Counterfactual-CAM, to embed the counterfactual representation perturbation mechanism into the vanilla CAM-based model. This mechanism efficiently decouples foreground and co-occurring contexts while synthesizing counterfactual representations.
  \item Extensive experiments conducted on multiple benchmark datasets demonstrate that Counterfactual-CAM successfully mitigates the ``biased activation'' problem and achieves remarkable improvements over prior state-of-the-art approaches.
\end{itemize}

\section{Related Work}
\subsection{Weakly-supervised Object Localization}
To address the WSOL task, the most common solutions have relied on class activation maps (CAMs)~\cite{zhou2016learning, shao2022deep} to segment the highest activation area as a coarse object localization. Prevailing works in this vein~\cite{kim2017two, zhang2018adversarial, babar2021look} try to solve the most discriminative region localization problem in vanilla CAM. To overcome the problem, the community has developed several methods~\cite{diba2017weakly, guo2021strengthen, wei2021shallow, kim2017two, zhang2018adversarial, babar2021look}  that aim to perceive the entire object rather than the contracted and sparse discriminative regions. One category of methods~\cite{li2016weakly, wei2018ts2c} addresses this issue by selecting positive proposals based on the discrepancy between their information and that of their surrounding contextual regions. WSLPDA~\cite{li2016weakly} and TS$^2$C~\cite{wei2018ts2c} compare pixel values within a proposal and its neighboring contextual region. Another approach involves the use of a cascaded network structure~\cite{diba2017weakly, guo2021strengthen, wei2021shallow} to expand and refine the initial prediction box. The output from the preceding stage acts as the pseudo ground-truth to supervise subsequent training stages. Furthermore, some methods, such as TP-WSL~\cite{kim2017two}, ACoL~\cite{zhang2018adversarial}, ADL~\cite{choe2019attention}, and MEIL~\cite{mai2020erasing} adopt an erasing strategy to compel the detector to identify subsequent discriminative regions. Additionally, SPOL~\cite{wei2021shallow} and ORNet~\cite{xie2021online} leverage the low-level feature to preserve more object detail information in the object localization phase.

In contrast to the extensively explored "most discriminative region localization" problem, the issue of "biased activation" induced by co-occurring background has received less attention. This paper endeavors to tackle co-occurring backgrounds via counterfactual learning.

\subsection{Causal Inference}
Causal intervention as the effective solution in addressing confounder problem is widely used in various tasks, such as few-shot learning~\cite{yue2020interventional}, long-tailed classification~\cite{tang2020long}, and weakly-supervised segmentation~\cite{zhang2020causal} and localization~\cite{shao2021improving}. Taking weakly-supervised segmentation and localization for example, without the instance- and pixel-label supervision, the context as a confounding factor leads image-level classification models to learn spurious correlations between pixels and labels. To solve them, CONTA~\cite{zhang2020causal} and CI-CAM~\cite{shao2021improving} introduce the context adjustment based on backdoor adjustment~\cite{pearl2016causal} to remove the effect of context on the image. 
Counterfactual analysis can handle unmeasured or unknown confounding factors widely used in various tasks. For example, TDE~\cite{tang2020unbiased} uses the counterfactual causality to infer the effect from bad bias and uses the total direct effect to achieve unbiased prediction. Chen \etal~\cite{chen2020counterfactual} generates counterfactual samples by masking critical objects in images or words in questions to reduce the language biases in VQA. CAL~\cite{rao2021counterfactual} leverages counterfactual analysis to fine-grained visual categorization and re-identification by maximizing the prediction of original and counterfactual attention. FairCL~\cite{zhang2023fairnessaware} generates counterfactual images for self-supervised contrastive learning to improve the fairness of learned representations. 

In our work, we simulate counterfactual scenarios by pairing the constant foreground with various backgrounds. By training the model with these counterfactual scenarios using the correct label, we can naturally lead the model to focus on the constant foreground content while disregarding the varying background information. To our knowledge, this work represents the first attempt to utilize counterfactual learning in this direction.

\begin{figure*}[t]
  \centering
  \includegraphics[width=0.9\linewidth]{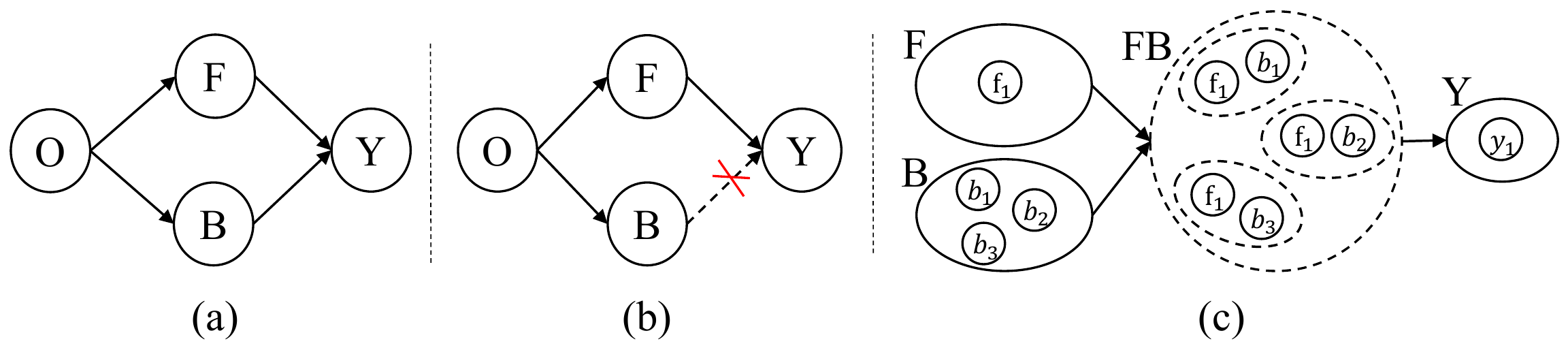}
  % \vspace{-0.5em}
  \caption{(a) Building the structural causal model (SCM) in WSOL. (b) Cutting off the confounding effect of $B \rightarrow Y$ in WSOL. (c) Synthesizing counterfactual representations to remove the confounding effect of $B \rightarrow Y$. $O$: original image feature. $F$: foreground feature, $f_1 \in F$. $B$: background feature, $b_1, b_2, b_3 \in B$. $FB$: synthesized counterfactual representation. $Y$: image label, $y_1 \in Y$.}
  \label{structural_causal_model_graph}
  \vspace{-1.0em}
\end{figure*} 
\section{Methodology}

\subsection{Preliminaries}
\subsubsection{Combinational Class Activation Mapping}
Given an image $I$, we extract feature maps $X \in \mathbb{R}^{c \times h \times w}$ using a backbone network, where $c$, $h$, and $w$ denote channel count, height, and width. 
A global average pooling (GAP) layer transforms $X$ into a feature vector $V \in \mathbb{R}^{c}$. 
A classifier with a weight matrix $W \in \mathbb{R}^{n \times c}$, where $n$ is the class count, produces prediction class $i$ based on $V$. 
The activation map $M_i$ for class $i$ in class activation maps (CAMs) $M \in \mathbb{R}^{n \times h \times w}$ is then computed by:
\begin{equation}
  \begin{aligned}
     M_i = \sum_{j}^{c} {W_{i, j} \cdot X_j}, 1 \leq i \leq n.
  \end{aligned}
  \label{eq:cam}
\end{equation}

NL-CCAM~\cite{yang2020combinational} argues that the activation map $M_i$ of the prediction class $i$ often biases to over-small regions or sometimes even highlights background area. Thus, it proposes a combinational class activation mapping by combining all activation maps to generate a better localization map $H$. 
\begin{equation}
   \begin{aligned}
      H = \sum_{i}^n \omega_i\cdot M_i,
   \end{aligned}
   \label{eq:ccam}
\end{equation}
where $\omega_i$ is a combinational weight associated with the ranked index of class $i$. In this work, we build upon NL-CCAM~\cite{yang2020combinational} but introduce significant improvement. Specifically, We equip the baseline with the ability to solve the ``biased activation'' problem. 

\subsubsection{Structural Causal Model}
Inspired by CONTA~\cite{zhang2020causal}, a structural causal model~\cite{pearl2016causal} is utilized to analyze the causal relationship among original image feature $O$, foreground feature $F$, background feature $B$, and image label $Y$. The direct link shown in Figure~\ref{structural_causal_model_graph} (a) represents the causality from one node to another, indicating a cause $\rightarrow$ effect relationship~\cite{zhang2020causal}.

$\bm{F \leftarrow O \rightarrow B}$: It indicates that the original image feature $O$ consists of the foreground feature $F$ and the background feature $B$. For example, in an image of a fish, the foreground corresponds to the ``fish'' and the background corresponds to the ``water''.

$\bm{F \rightarrow Y \leftarrow B}$: It denotes that the image prediction $Y$ of the original image is affected by both the foreground feature $F$ and the background feature $B$. However, without instance-level labels, the model inspection makes it hard to distinguish between the foreground and its co-occurring background, resulting in the wrong activation. For instance, in cases where the background, such as ``water," is mistakenly activated as the foreground ``fish", or when a ``bird" drinking by the river is erroneously classified as a ``fish" due to the improper activation of the ``water."

To remove the negative effect from the background ``water'' and cut off the link of $B \rightarrow Y$, we construct a complete event (\eg, background) group. Specifically, we pair the foreground with all of backgrounds and assign the foreground category to these synthesized representations as shown in Figure~\ref{structural_causal_model_graph} (c). Following the total probability formula in Equation~\ref{eq:scm}, we obtain a pure prediction between $F$ and $Y$. 
\begin{equation}
  \begin{aligned}
    P(Y|F) & = \sum_{i}^{D} P(Y|F, b_i)\cdot P(b_i|F),
  \end{aligned}
  \label{eq:scm}
\end{equation}
where $D$ and $B = \{b_1, b_2, ..., b_D\}$ are the number of training images and a comprehensive background set, respectively. The assumption of independence between foreground $F$ and background $B$ allows replacing $P(b_i|F)$ with $P(b_i)$. Given that the occurrence probability of images in the dataset is roughly uniform, we set $P(b_i)$ to $1/D$. Consequently, $P(Y|F)$ can be expressed as $1/D\cdot\sum_{i}^{D} P(Y|F, b_i)$.

\subsection{Technical Details of Counterfactual-CAM}
\begin{figure*}[t]
  \centering
 \includegraphics[width=1.0\linewidth]{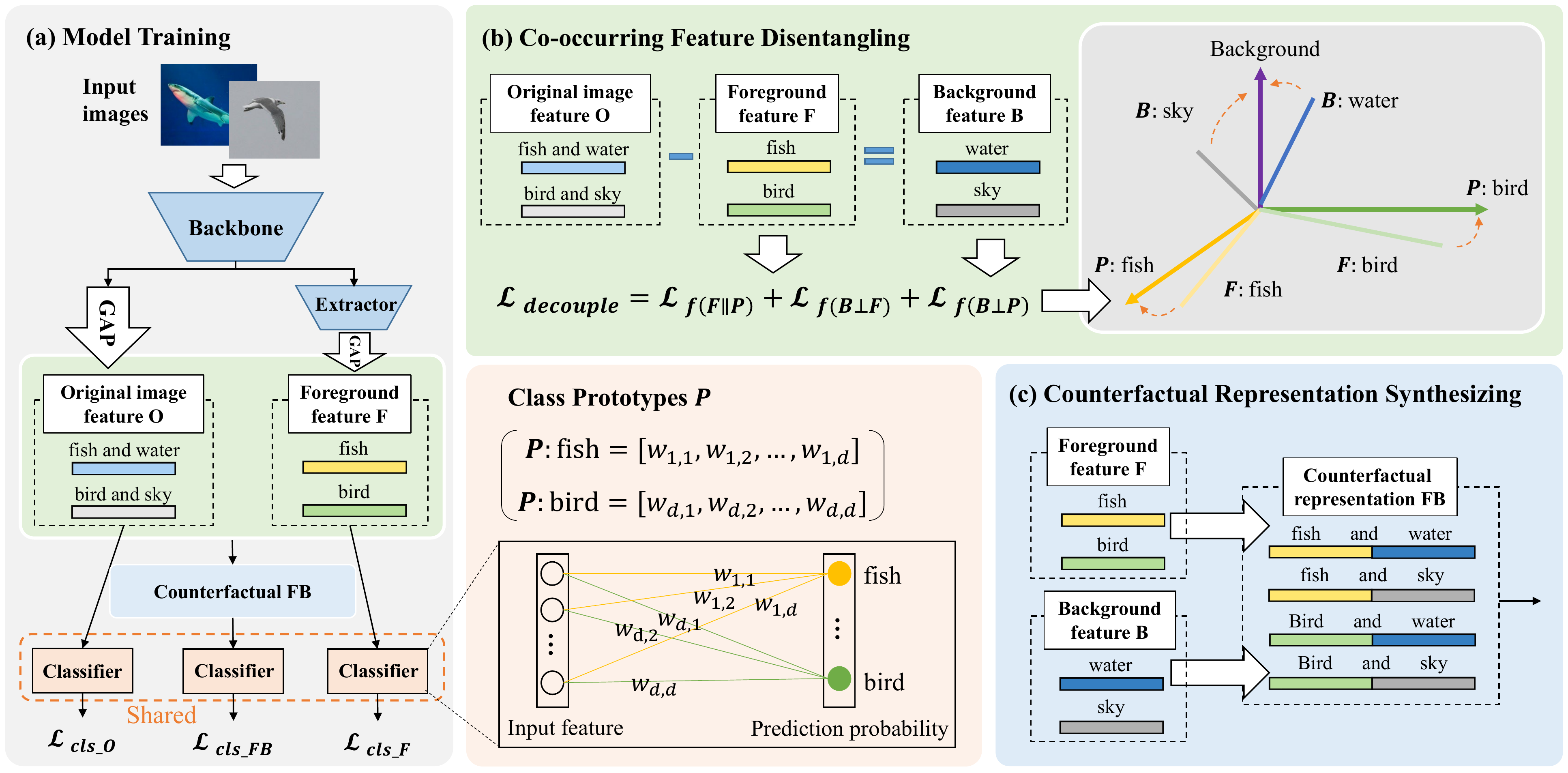}
%  \vspace{-0.5em}
  \caption{Overview of the proposed Counterfactual-CAM. (a) The learning process of Counterfactual-CAM. $d$ denotes the length of the prototype feature. (b) Decoupling original image feature to foreground feature and background feature. (c) Synthesizing counterfactual representations by pairing each foreground feature and various background features.}
  \label{main}
\vspace{-1.0em}
\end{figure*}

To address the issue of ``biased activation" as outlined in Equation~\ref{eq:scm}, we introduce a counterfactual model referred to as Counterfactual-CAM. 
Its key components include a counterfactual representation perturbation mechanism, which comprises feature disentangling for co-occurring elements and the synthesis of counterfactual representations.

\subsubsection{Co-occurring Feature Disentangling}
Given an image $I$, we first obtain its feature maps $X$ through a backbone network. 
Then, $X$ is fed into a global average pooling (GAP) layer to produce the original feature $O \in \mathbb{R}^{d}$, where $d$ is the feature dimension. Meanwhile, $X$ is forwarded into a foreground extractor (\ie, two convolutional layers) and a GAP layer to generate image foreground feature $F \in \mathbb{R}^{d}$. 
Finally, we can separate background feature $B \in \mathbb{R}^{d}$ from original feature $O$ by subtracting foreground feature $F$ as shown in Figure~\ref{main} (b). 
To ensure the accuracy of the decoupling process, we set up the following two rules: 

\textbf{Rule 1.} Foreground feature $F$ should be parallel with its corresponding class prototype $P$:
\begin{equation}
  \begin{aligned}
    \mathcal{L}_{f(F \parallel P)} = \sum_{i \neq k}^{n} L_1(\frac{F \cdot p_i}{|F| \times |p_i|}, 0) 
 + L_1(\frac{F \cdot p_k}{|F|\times |p_k|}, 1),
  \end{aligned}
  \label{eq:parallel}
\end{equation}
where $L_1$, $n$, and $p_k$ indicate the L1 distance function, the number of classes, and the class prototype of class $k$, respectively. Inspired by T3A~\cite{iwasawa2021test} and PCT~\cite{tanwisuth2021prototype}, we use the weight of the classifier as class prototypes $P=\{p_1, p_2, ..., p_n\}$. Equation~\ref{eq:parallel} aims to align the foreground feature $F$ with its corresponding class prototype $p_k$. 

\textbf{Rule 2.} Background feature $B$ should be orthogonal from all foreground feature $F$ and class prototype $P$:
\begin{equation}
  \begin{aligned}
   \mathcal{L}_{f(B\perp F)} = L_1(\frac{B \cdot F}{|B| \times |F|}, 0), \\ 
   \mathcal{L}_{f(B\perp P)} = \sum_{i}^{n} L_1(\frac{B \cdot p_i}{|B| \times |p_i|}, 0),
  \end{aligned}
  \label{eq:orthogonal}
\end{equation}
where $f(B \perp F)$ and $f(B \perp P)$ are optimal feature orthogonal strategy between the background feature $B$ with all foreground features $F$ and class prototypes $P$. 

Building upon the two principles, we introduce a decoupled loss, denoted as $\mathcal{L}_{decouple}$, to efficiently disentangle the foreground feature $F$ and background feature $B$ from the original feature $O$. 
The loss is formulated as follows:
\begin{equation}
  \begin{aligned}
    \mathcal{L}_{decouple} = \mathcal{L}_{f(F \parallel P)} + \mathcal{L}_{f(B \perp F)} + \mathcal{L}_{f(B \perp P)}.
  \end{aligned}
  \label{eq:decouple}
\end{equation}

Taking Figure~\ref{main} (b) as an example, $f(F \parallel P)$ aligns the feature of foreground, such as ``fish"($F: fish$), with its corresponding class prototype, ``fish"($P: fish$). In contrast, $f(B \perp F)$ and $f(B \perp P)$ aim to disentangle background features, such as ``water" and ``sky", from foreground features and class prototypes. 
Our design ensures that the foreground and background features are effectively separated.

\subsubsection{Counterfactual Representation Synthesizing}
To remove the confounding effect of $B \rightarrow Y$ as shown in Figure~\ref{structural_causal_model_graph} (b), we intend to leverage the total probability formula to pursue the pure causality between the cause $F$ and the effect $Y$ as shown in Equation~\ref{eq:scm}. Specifically, we first collect all the background features $B = \{b_1, b_2, ...\}$. Then, we pair each foreground feature $F$ with various background features $B = \{b_1, b_2, ...\}$ to synthesize a large number of counterfactual representations $FB$ as shown in Figure~\ref{structural_causal_model_graph} (c). Finally, we assign a label to each counterfactual representation $FB$ according to its foreground category. 

Taking Figure~\ref{main} (c) for example, we have the foreground ``fish'' and ``bird'' as well as background ``water'' and ``sky''. After coupling foregrounds and backgrounds, we generate four synthesized representations: ``fish and water'', ``fish and sky'', ``bird and water'', and ``bird and sky''. Therein, ``fish and sky'' and `bird and water'' are counterfactual representations. By aligning the prediction between the original image representations (\ie, ``fish and water'') and counterfactual representations (\ie, ``fish and sky''), we compel the CAM-based model to focus on the constant foreground ``fish'' while disregarding the ``sky'' and ``water'' information (likewise for ``bird'').

\subsubsection{Training Objective}
Our proposed network not only learns to optimize the classification losses of the original image, foreground, and counterfactual representation but also learns to minimize the decoupled loss $\mathcal{L}_{decouple}$ to ensure the accuracy of the co-occurring feature disentangling. Given an image, we first obtain the original image prediction score $s^o$, foreground prediction score $s^f$, and counterfactual representation prediction score $s^{fb}$. Then we train Counterfactual-CAM using the following loss function $\mathcal{L}_{train}$.
\begin{equation}
   \begin{aligned}
      \mathcal{L}_{train} & =\mathcal{L}_{ce}(s^o, y)+\mathcal{L}_{ce}(s^f, y)+\mathcal{L}_{ce}(s^{fb}, y) \\
      & +\alpha \cdot \mathcal{L}_{decouple},
   \end{aligned}
   \label{eq:train_loss}
\end{equation}
where $\mathcal{L}_{ce}$, $y$, and $\alpha$ respectively denote the cross entropy function, image label, and hyperparameter. 

\begin{figure*}[t]
  \centering
  \includegraphics[width=1.0\linewidth]{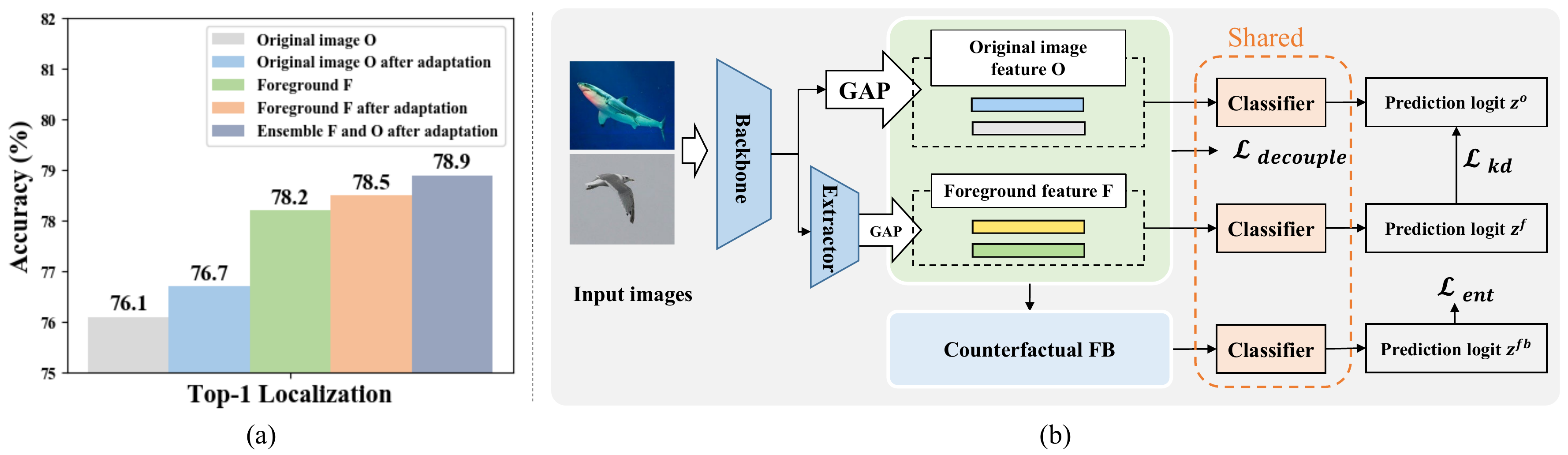}
  % \vspace{-0.5em}
  \caption{(a) Comparison of the prediction of the original image $O$, foreground $F$, and adaptation. (b) Overview of test-time adaptation, which finetunes the BN layers, foreground extractor, and classifier.}
  \label{test_time}
  \vspace{-1.0em}
\end{figure*}

\subsection{Test-time Counterfactual Adaptation}
The training and testing set usually suffer from a distribution gap on their co-occurring backgrounds, which hinders CAM from highlighting the accurate objects. To fully leverage the foreground hints present in test images to boost our CCL performance, we draw inspiration from the design of tent~\cite{wang2020tent} and propose an online adaptation strategy with the following two considerations:

\textbf{Consideration 1:} The information from the test images provides valuable insights into the specific objects and their context present in the input images.

\textbf{Consideration 2:} Feeding the test-set foreground information into the detection model helps to activate the object foreground region and suppress the background region.

More concretely, for \textbf{Consideration 1}, we first aim to use the $\mathcal{L}_{decouple}$ (cf. Equation~\ref{eq:decouple}) to thoroughly decouple foreground and background from the original image. Then, minimizing the Shannon Entropy upon the prediction of the counterfactual representation to further align the constant foreground of the counterfactual representation and its corresponding class prototype. For \textbf{Consideration 2}, we take foreground knowledge to distill the original image prediction to force the model to pay more attention to the foreground. Finally, the total adaptation loss $\mathcal{L}_{adapt}$ is given as follows.
\begin{equation}
  \begin{aligned}
   &\mathcal{L}_{adapt} = \beta \cdot \mathcal{L}_{kd}
   + (1-\beta) \cdot (\mathcal{L}_{ent}(z^{fb}) 
    + \delta \cdot
    \mathcal{L}_{decouple}), \\
   &\mathcal{L}_{kd} = KL(\frac{\exp(z^{o}/T)}{\sum_{j=1}^{n} \exp(z^{o}_j/T)}, \frac{\exp(z^{f}/T)}{\sum_{j=1}^{n} \exp(z^{f}_j/T)}),
  \end{aligned}
  \label{eq:adaptation}
\end{equation}
where $\mathcal{L}_{ent}$, $KL$, $z^o$, $z^f$, and $z^{fb}$ denote the Entropy loss, Kullback-Leibler divergence loss, original image logit, foreground logit, and counterfactual representation logit, respectively. $T$ and $n$ respectively denote the distillation temperature and class number. $\beta$ and $\delta$ are the hyperparameters.

\section{Experiments}
\subsection{Experimental Settings}
% \textbf{Datasets.} We use CUB-200-2011~\cite{wah2011caltech}, ILSVRC 2016~\cite{russakovsky2015imagenet} and OpenImages30k~\cite{choe2020evaluating} as experimental datasets.
% \textbf{Evaluation Metrics.} We utilize Accuracy, MaxBoxAccV2~\cite{choe2020evaluating}, and PxAP~\cite{choe2020evaluating} as our primary evaluation metrics. Accuracy includes Top-1 classification (Top-1 Cls), Top-1 localization (Top-1 Loc), and GT-known localization (GT-known) accuracy. 
% \textbf{Implementation Details.} We adopted the VGG16~\cite{simonyan2014very}, InceptionV3~\cite{szegedy2016rethinking}, and Resnet50~\cite{he2016deep} pre-trained on the ImageNet~\cite{russakovsky2015imagenet} as our backbones. \footnote{For detailed Dataset introduction, Evaluation Metric introduction, and Implementation Details, please refer to the appendix.}

\textbf{Datasets.} The evaluation of the proposed Counterfactual-CAM was conducted on three datasets: CUB-200-2011~\cite{wah2011caltech}, ILSVRC 2016~\cite{russakovsky2015imagenet}, and OpenImages30k~\cite{choe2020evaluating}. \textit{1) CUB-200-2011~\cite{wah2011caltech}} focuses on subordinate categorization, comprising $200$ bird categories. It includes $5,994$ images with image-level labels in the training set and $5,794$ images with instance-level labels in the test set. \textit{2) ILSVRC 2016~\cite{russakovsky2015imagenet}} contains $1,000$ categories and encompasses over $1.2$ million images with image-level labels in the training set and $50,000$ images with instance-level labels in the validation set. \textit{3) OpenImages30k~\cite{choe2020evaluating}} contains $100$ classes and comprises three disjoint sub-datasets: train-weaksup, train-fullsup, and test. Train-weaksup includes $29,819$ images with image-level labels, train-fullsup contains $2,500$ images with full supervision (either bounding box or binary mask), and the test set includes $5,000$ images with full supervision.

\textbf{Evaluation Metrics.} We utilize Accuracy, MaxBoxAccV2~\cite{choe2020evaluating}, and PxAP~\cite{choe2020evaluating} as our primary evaluation metrics. \textit{1) Accuracy} includes Top-1 classification accuracy (Top-1 Cls), Top-1 localization accuracy (Top-1 Loc), and GT-known localization accuracy (GT-known). Top-1 Cls assesses the accuracy of the highest prediction score being correct, while Top-1 Loc measures both category prediction and box localization accuracy. GT-known focuses solely on box localization prediction precision. \textit{2) MaxBoxAccV2~\cite{choe2020evaluating}} offers a more comprehensive evaluation by averaging performance across three Intersection over Union (IoU) thresholds (\eg, $0.3$, $0.5$, $0.7$) to cater to different localization precision requirements. \textit{3) Pixel Average Precision (PxAP)~\cite{choe2020evaluating}} evaluates the performance of prediction masks by calculating the area under the curve of the pixel precision-recall curve.

% \textbf{Implementation Details}. Please refer to the appendix.

\subsection{Implementation Details}
We utilize VGG16~\cite{simonyan2014very}, InceptionV3~\cite{szegedy2016rethinking}, and Resnet50~\cite{he2016deep} pretrained on ImageNet~\cite{russakovsky2015imagenet} as the backbone for our proposed Counterfactual-CAM. Additionally, the foreground extractor introduced in our Counterfactual-CAM consists of two convolutional layers and two activation functions. Notably, we apply RandAugment~\cite{szegedy2016rethinking} for data augmentation on the CUB-200-2011~\cite{wah2011caltech} dataset during training.

For the VGG16~\cite{simonyan2014very} backbone, we fine-tune our proposed Counterfactual-CAM with an Adam~\cite{kingma2014adam} optimizer by randomly cropping the input images to $224 \times 224$. In experiments on the CUB-200-2011~\cite{wah2011caltech} dataset, the learning rate is initially set to $0.0005$ and decays with a polynomial scheduler for later epochs until training reaches $100$ epochs. The batch size and the hyperparameter $\alpha$ of the train loss are set as $12$ and $0.001$. The hyperparameters $\beta$, $\delta$, and $T$ of the test-time counterfactual adaptation are set to $0.2$, $0.012$, and $15$, respectively. Similarly, on the ILSVRC 2016~\cite{russakovsky2015imagenet} dataset, the learning rate is initially set to $0.0000585$ and decays with a polynomial scheduler for later epochs until training reaches $20$ epochs. The batch size and the hyperparameter $\alpha$ of the train loss are set as $72$ and $0.12$. During testing, we resize the input images to $344 \times 344$ on the CUB-200-2011~\cite{wah2011caltech} dataset (\ie, $306 \times 306$ on the ILSVRC 2016~\cite{russakovsky2015imagenet} dataset) and perform a central crop of $244 \times 244$, inspired by~\cite{wu2022background, wei2021shallow, zhang2020rethinking, choe2019attention, yun2019cutmix}. Finally, we set the segmentation threshold to $0.14$ and $0.16$ for generating bounding boxes on the CUB-200-2011 and ILSVRC 2016 datasets, respectively.

For the InceptionV3~\cite{szegedy2016rethinking} backbone, we fine-tune our proposed Counterfactual-CAM with an Adam~\cite{kingma2014adam} optimizer by randomly cropping the input images to $299 \times 299$. In experiments on the CUB-200-2011~\cite{wah2011caltech} dataset, the learning rate is initially set to $0.0001$ and decays with a polynomial scheduler for later epochs until training reaches $100$ epochs. The batch size and the hyperparameter $\alpha$ of the train loss are set to $12$ and $0.0001$, respectively. The hyperparameters $\beta$, $\delta$, and $T$ of the test-time counterfactual adaptation are set to $0.1$, $0.012$, and $30$, respectively. Similarly, on the ILSVRC 2016~\cite{russakovsky2015imagenet} dataset, the learning rate is initially set to $0.0002$ and decays with a polynomial scheduler for later epochs until training reaches $20$ epochs. The batch size and the hyperparameter $\alpha$ of the train loss are set to $126$ and $1e-05$, respectively. During the testing phase, we resize the input images to $500 \times 500$ on the CUB-200-2011~\cite{wah2011caltech} dataset (i.e., $424 \times 424$ on the ILSVRC 2016~\cite{russakovsky2015imagenet} dataset) and perform a central crop of $299 \times 299$, inspired by~\cite{wu2022background, wei2021shallow, zhang2020rethinking, choe2019attention, yun2019cutmix}. Finally, we set the segmentation threshold to $0.21$ and $0.18$ for generating bounding boxes on the CUB-200-2011 and ILSVRC 2016 datasets, respectively.

For the Resnet50~\cite{he2016deep} backbone, we fine-tune our proposed Counterfactual-CAM using an Adam~\cite{kingma2014adam} optimizer, with random cropping of input images to $244 \times 244$. The learning rate is initially set to $5e-05$ and decays with a polynomial scheduler in subsequent epochs until training completes at $100$ epochs. On the CUB-200-2011~\cite{wah2011caltech} dataset, the batch size and the hyperparameter $\alpha$ of the training loss are set to $12$ and $0.012$, respectively. Similarly, on the ILSVRC 2016~\cite{russakovsky2015imagenet} dataset, the learning rate is initially set to $1e-5$ and decays with a polynomial scheduler for later epochs until training reaches $20$ epochs. The batch size and the hyperparameter $\alpha$ of the train loss are set to $72$ and $1e-04$, respectively. During the testing phase, we resize the input images to $344 \times 344$ on the CUB-200-2011~\cite{wah2011caltech} dataset (i.e., $280 \times 280$ on the ILSVRC 2016~\cite{russakovsky2015imagenet} dataset) and perform a central crop of $224 \times 224$, inspired by~\cite{wu2022background, wei2021shallow, zhang2020rethinking, choe2019attention, yun2019cutmix}. Finally, we set the segmentation threshold to $0.16$ and $0.22$ for generating bounding boxes on the CUB-200-2011 and ILSVRC 2016 datasets, respectively.

\subsection{Comparisons with State-of-The-Art Methods}

\addtolength{\tabcolsep}{-0.5pt} 
\begin{table}[t]
   \centering
   \caption{Comparison with the state-of-the-art methods on the CUB-200-2011 dataset.}
   \label{tab:sota_acc_cub}
   \begin{tabular}{lcccc}
   \toprule
   % \multicolumn{1}{c|}{\multirow{2}{*}{Method}} & \multicolumn{1}{c|}{\multirow{2}{*}{Backbone}} & \multicolumn{3}{c}{CUB-200-2011}                                   \\
   % \multicolumn{1}{c|}{} &   \multicolumn{1}{c|}{} & Top-1 Cls & Top-1 Loc & GT-known   \\ \hline
   Method & Backbone & Top-1 Cls & Top-1 Loc & GT-known\\
   \midrule
   NL-CCAM~\cite{yang2020combinational}  &VGG16    &  73.4        &    52.4  & -  \\ 
   MEIL~\cite{mai2020erasing}   &VGG16        &  74.8       &    57.5  & 73.8   \\
   PSOL~\cite{zhang2020rethinking}&VGG16 & - & 66.3 & -  \\
   GCNet~\cite{lu2020geometry}&VGG16  & 76.8 & 63.2 & - \\
   RCAM~\cite{bae2020rethinking}&VGG16   &  74.9        &     61.3   & 80.7  \\
   MCIR~\cite{babar2021look}&VGG16 & 72.6 & 58.1 & - \\
   SLT-Net~\cite{guo2021strengthen}&VGG16  & 76.6 & 67.8& 87.6 \\
   SPA~\cite{pan2021unveiling}&VGG16 & - & 60.3 & 77.3 \\
   ORNet~\cite{xie2021online}&VGG16  & 77.0& 67.7 & - \\
   % FAM~\cite{meng2021foreground}&VGG16  & \blue{77.3} & 69.3 & 89.3 &  70.9 & 52.0 & \textbf{71.7}\\
   % PDM~\cite{meng2022diverse}&VGG16  & 76.9 & 67.3 & 82.2 &  68.7 & 51.1 & \blue{69.3}\\
   % BAS~\cite{wu2022background}&VGG16 & - & \blue{71.3} & 91.1 &  - & \red{53.0} & \blue{69.6}  \\
   BridgeGap~\cite{kim2022bridging}&VGG16 & - & \blue{70.8} & \textbf{93.2}   \\
   CREAM~\cite{xu2022cream}&VGG16 & - & 70.4 & 91.0 \\
   \rowcolor{mygray}Ours&VGG16 & \textbf{79.4} &	\textbf{74.1} &	92.8 \\ 
   \hline
   PSOL~\cite{zhang2020rethinking}&InceptionV3 &-&65.5&-\\
   % MEIL~\cite{mai2020erasing}&IncepV3 &-&-&-\\
   GC-Net~\cite{lu2020geometry}&InceptionV3 &-&58.6&75.3\\	
   I$^2$C~\cite{zhang2020inter}&InceptionV3 &-&55&72.6\\
   SLT-Net~\cite{guo2021strengthen}&InceptionV3 &76.4&66.1&86.5\\
   SPA~\cite{pan2021unveiling}&InceptionV3 &-&53.6&72.1\\
   % FAM~\cite{meng2021foreground}&IncepV3 &\textbf{81.3}&70.7&87.3&\blue{77.6}&55.2&68.6\\
   % PDM~\cite{meng2022diverse}&IncepV3 &-&64.3&79.6&-&-&-\\			
   % BAS~\cite{wu2022background}&IncepV3 &-&\blue{73.3}&\blue{92.2}&-&58.5&71.9\\
   CREAM~\cite{xu2022cream}&InceptionV3 &-&\blue{71.8}&\blue{90.4}\\
   \rowcolor{mygray}Ours&InceptionV3 &\textbf{82.2} & \textbf{78.3}	& \textbf{95.0} \\
   \hline
   PSOL~\cite{zhang2020rethinking}&Resnet50 &-&70.7&\blue{90.0}\\
   RCAM~\cite{bae2020rethinking}&Resnet50 &75.0&59.5&77.6\\
   % FAM~\cite{meng2021foreground}&Resnet &\blue{82.7}&73.7&85.7&\textbf{76.5}&\blue{54.5}&64.6\\
   % PDM~\cite{meng2022diverse}&Resnet &81.3&71.2&82.3&75.6&54.4&\textbf{69.6}\\
   % BAS~\cite{wu2022background}&Resnet &-&\blue{77.3}&\blue{95.1}&-&57.2&71.8\\
   CREAM~\cite{xu2022cream}&Resnet50 &-&\blue{76.0}&89.9\\
   % \rowcolor{mygray}KG-CI-CAM&Resnet &\textbf{84.0}&\textbf{81.6}&\textbf{96.8}& 75.0 & 54.2 &68.9\\	
   \rowcolor{mygray}Ours&Resnet50 &\textbf{84.9}&\textbf{81.7}&\textbf{96.0}\\	
   \hline

   \hline
   \end{tabular}
   \vspace{-1.0em}
\end{table}
\addtolength{\tabcolsep}{0.5pt} 

\addtolength{\tabcolsep}{+1.5pt} 
\begin{table}[t] %[!t]
  \centering
  % \caption{Performance using MaxBoxAccV2 as the evaluation metric on the CUB dataset.}
  \caption{MaxBoxAccV2 on the CUB-200-2011 dataset.}
  % \vspace{-1.0em}
  \label{tab:sota_accV2_cub}
  \begin{tabular}{lcccc}
  \toprule
  Method            & VGG16 & InceptionV3 & Resnet50 & Mean \\ 
  \midrule
  CAM~\cite{zhou2016learning}       & 63.7  & 56.7        & 63.0       & 61.1 \\ 
  HaS~\cite{keeler1990integrated}      & 63.7  & 53.4        & 64.7     & 60.6 \\ 
  ACoL~\cite{zhang2018adversarial}     & 57.4  & 56.2        & 66.5     & 60.0   \\ 
  SPG~\cite{papandreou2015weakly}      & 56.3  & 55.9        & 60.4     & 57.5 \\ 
  ADL~\cite{choe2019attention}        & 66.3  & 58.8        & 58.4     & 61.0   \\ 
  CutMix~\cite{joon2017exploiting}   & 62.3  & 57.5        & 62.8     & 58.8 \\ 
  CAM\_IVR~\cite{girshick2014rich} & 65.2  & 60.8        & 66.9     & 64.2 \\ 
  CREAM~\cite{xu2022cream}    & \textbf{71.5}  & 64.2        & 73.5     & 69.7 \\ 
  \rowcolor{mygray}
  Ours & 66.6  & \textbf{67.4}        & \textbf{76.5}     & \textbf{70.2} \\
  \bottomrule 
  \end{tabular}
  % \vspace{-1.0em}
\end{table}
\addtolength{\tabcolsep}{-1.5pt} 

We compared Counterfactual-CAM with other state-of-the-art (SOTA) methods on the CUB-200-2011~\cite{wah2011caltech}, ILSVRC 2016~\cite{russakovsky2015imagenet} and OpenImages30k~\cite{choe2020evaluating} datasets. The final experimental results of our method are the ensemble classification of the original image and foreground.

% The results of the first and second are shown in bold and blue, respectively. 
\addtolength{\tabcolsep}{-0.5pt} 
\begin{table}[t]
   \centering
   \caption{Comparison with the state-of-the-art methods on the ILSVRC 2016. $\ast$ indicates the re-implemented results of baseline model by ourselves.}
   % \vspace{-3mm}
   % \resizebox{2\columnwidth}{!}{
   \label{tab:sota_acc_imagenet}
   \begin{tabular}{lcccc}
   % \hline
   % \hline
   % \multicolumn{1}{c|}{\multirow{2}{*}{Method}} & \multicolumn{1}{c|}{\multirow{2}{*}{Backbone}} & \multicolumn{3}{c|}{CUB-200-2011}                                                    & \multicolumn{3}{c}{ILSVRC 2016}                                                                  \\
   % \multicolumn{1}{c|}{} &   \multicolumn{1}{c|}{} & Top-1 Cls & Top-1 Loc & GT-known        &  Top-1 Cls & Top-1 Loc & GT-known     \\ \hline
   \toprule
   Method & Backbone & Top-1 Cls & Top-1 Loc & GT-known\\
   \midrule
   % NL-CCAM~\cite{yang2020combinational}  &VGG16      &   72.3      &    50.2 & 65.2   \\ 
   NL-CCAM~\cite{yang2020combinational}$\ast$  &VGG16      &   72.3      &    48.6 & 62.9   \\
   MEIL~\cite{mai2020erasing}   &VGG16          &   70.3        &     46.8   & -   \\
   PSOL~\cite{zhang2020rethinking}&VGG16  &  - & 50.9 & 64.0 \\
   GCNet~\cite{lu2020geometry}&VGG16   &  - & - & - \\
   RCAM~\cite{bae2020rethinking}&VGG16     &   67.2        &      45.4   & 62.7\\
   MCIR~\cite{babar2021look}&VGG16  &  71.2 & 51.6 & 66.3\\
   SLT-Net~\cite{guo2021strengthen}&VGG16  &  \blue{72.4} & 51.2 & 67.2\\
   SPA~\cite{pan2021unveiling}&VGG16  &  - & 49.6 & 65.1\\
   ORNet~\cite{xie2021online}&VGG16   &  71.6 & \blue{52.1} & -\\
   % FAM~\cite{meng2021foreground}&VGG16  & \blue{77.3} & 69.3 & 89.3 &  70.9 & 52.0 & \textbf{71.7}\\
   % PDM~\cite{meng2022diverse}&VGG16 & 68.7 & 51.1 & \blue{69.3}\\
   % BAS~\cite{wu2022background}&VGG16 & - & \blue{71.3} & 91.1 &  - & \red{53.0} & \blue{69.6}  \\
   BridgeGap~\cite{kim2022bridging}&VGG16   &  - & 49.9 & \textbf{68.9} \\
   CREAM~\cite{xu2022cream}&VGG16  & - & \textbf{52.4} & 68.3\\
  %  \rowcolor{mygray}Ours&VGG16  & \textbf{72.8}	& 52.1 & 68.1 \\ 
   \rowcolor{mygray}Ours&VGG16  & \textbf{72.5}	& \blue{52.1} & \blue{68.4} \\ 
   \hline
   NL-CCAM~\cite{yang2020combinational}$\ast$ &InceptionV3      &   73.1      &    52.9 & 66.8   \\
   PSOL~\cite{zhang2020rethinking}&InceptionV3 &-&54.8&65.2\\
   MEIL~\cite{mai2020erasing}&InceptionV3 &73.3&49.5&-\\
   GC-Net~\cite{lu2020geometry}&InceptionV3 &\blue{77.4}&49.1&-\\	
   I$^2$C~\cite{zhang2020inter}&InceptionV3&73.3&53.1&68.5\\
   SLT-Net~\cite{guo2021strengthen}&InceptionV3 &\textbf{78.1}&\blue{55.7}&67.6\\
   SPA~\cite{pan2021unveiling}&InceptionV3 &-&52.7&68.3\\
   % FAM~\cite{meng2021foreground}&IncepV3 &\textbf{81.3}&70.7&87.3&\blue{77.6}&55.2&68.6\\
   % PDM~\cite{meng2022diverse}&IncepV3 &-&64.3&79.6&-&-&-\\			
   % BAS~\cite{wu2022background}&IncepV3 &-&\blue{73.3}&\blue{92.2}&-&58.5&71.9\\
   CREAM~\cite{xu2022cream}&InceptionV3 &-&\textbf{56.1}&\blue{69.0}\\
   \rowcolor{mygray}Ours&InceptionV3 &   73.5	& 55.0 & \textbf{71.5} \\ 
   \hline
   NL-CCAM~\cite{yang2020combinational}$\ast$  &Resnet50      &   \blue{74.2}      &    53.2 & \blue{68.3}   \\
   PSOL~\cite{zhang2020rethinking}&Resnet50 &-&54.0&65.4\\
   RCAM~\cite{bae2020rethinking}&Resnet50 &\textbf{75.8}&49.4&62.2\\
   % FAM~\cite{meng2021foreground}&Resnet50 &\blue{82.7}&73.7&85.7&\textbf{76.5}&\blue{54.5}&64.6\\
   % PDM~\cite{meng2022diverse}&Resnet50 &75.6&54.4&69.6\\
   % BAS~\cite{wu2022background}&Resnet50 &-&\blue{77.3}&\blue{95.1}&-&57.2&71.8\\
   CREAM~\cite{xu2022cream}&Resnet50 &-&\textbf{55.7}&\textbf{69.3}\\
  \rowcolor{mygray}Ours&Resnet50 & \textbf{75.8} & \blue{54.9} &\textbf{69.3}\\
  \hline

   \hline
   \end{tabular}
   \vspace{-1.0em}
\end{table}
\addtolength{\tabcolsep}{0.5pt} 

For the simple scenarios as on the CUB-200-2011 whose background only consists of ``water'', ``sky'', ``tree'', ``grassland'' etc, Counterfactual-CAM significantly outperforms current state-of-the-art (SOTA) methods. Referencing Table~\ref{tab:sota_acc_cub} and Table~\ref{tab:sota_accV2_cub}, Counterfactual-CAM consistently arrives the best Top-1 Cls and Top-1 Loc accuracy across various backbones. For InceptionV3~\cite{szegedy2016rethinking} and Resnet50~\cite{he2016deep} backbones, Counterfactual-CAM attains the highest GT-known accuracy and MaxBoxAccV2~\cite{choe2020evaluating}. Although lagging behind GT-known SOTA method BridgeGap~\cite{kim2022bridging} by $0.4\%$ in GT-known accuracy, Counterfactual-CAM exhibits a remarkable $3.3\%$ improvement over BridgeGap~\cite{kim2022bridging} in Top-1 Loc when using VGG16~\cite{simonyan2014very} as the backbone. Despite trailing the MaxBoxAccV2 SOTA method CREAM~\cite{xu2022cream} with VGG16 as the backbone, it demonstrates significant improvement over CREAM~\cite{xu2022cream} with InceptionV3 and Resnet50 backbones, achieving the highest mean MaxBoxAccV2 performance.

% by $4.9\%$ 

For more complex scenarios, such as the ILSVRC 2016 dataset, which exhibits diverse and intricate backgrounds, Counterfactual-CAM performs comparably to current SOTA methods in Table~\ref{tab:sota_acc_imagenet}. Firstly, if the backbone is VGG16, Counterfactual-CAM achieves a Top-1 Cls accuracy of $72.5\%$, surpassing the current SOTA SLT-Net~\cite{guo2021strengthen} by $0.1\%$, while respectively improving both Top-1 Loc and GT-known accuracy by $0.9\%$ and $1.2\%$. Despite a slight deficiency in Top-1 Loc and GT-known compared to other SOTA methods, given the challenges posed by the large-scale ILSVRC 2016 dataset with its diverse scenarios and backgrounds, Counterfactual-CAM still achieves the second best in localization performance. Secondly, with InceptionV3 as the backbone, Counterfactual-CAM achieves a GT-known accuracy of $71.5\%$, surpassing the current SOTA CREAM~\cite{xu2022cream} by $2.5\%$. Although there is a slight deficiency in Top-1 Cls and Top-1 Loc performance compared to other SOTA methods, Counterfactual-CAM still ranks among the top 3 in Top-1 Cls and Top-1 Loc. Thirdly, if the backbone is Resnet50, Counterfactual-CAM achieves the best Top-1 Cls and GT-known Loc performance. Meanwhile, Counterfactual-CAM reaches the second best in Top-1 Loc.

To assess the quality of feature disentanglement and foreground activation using ground truth masks, we conduct a complementary experiment using the PxAP~\cite{choe2020evaluating} metric on the OpenImages30k dataset, as presented in Table~\ref{tab:sota_pxap}. The results illustrate that Counterfactual-CAM achieves the best segmentation performance, vividly illustrating the positive impact of our method on successful feature disentanglement.

In summary, our experimental results underscore the effectiveness and robustness of Counterfactual-CAM across diverse datasets and backbone architectures, establishing its superiority over existing SOTA methods.

\addtolength{\tabcolsep}{15pt} 
\begin{table}[] %[!t]
  \centering
  \caption{Performance on the OpenImages30k dataset.}
  \label{tab:sota_pxap}
  % \vspace{-1.0em}
  \begin{tabular}{lcl}
  \toprule
  Method             & Backbone    & PxAP \\
  \midrule
  CAM~\cite{zhou2016learning}       & InceptionV3 & 63.2 \\
  HaS{\cite{keeler1990integrated}}       & InceptionV3 & 58.1 \\
  ACoL~\cite{zhang2018adversarial}      & InceptionV3 & 57.2 \\
  SPG~\cite{papandreou2015weakly}       & InceptionV3 & 62.3 \\
  ADL~\cite{choe2019attention}         & InceptionV3 & 56.8 \\
  CutMix~\cite{joon2017exploiting}    & InceptionV3 & 62.5 \\
  RCAM~\cite{bae2020rethinking}        & InceptionV3 & 63.3 \\
  CAM\_IVR~\cite{girshick2014rich}  & InceptionV3 & 63.6 \\
  CREAM~\cite{xu2022cream}     & InceptionV3 & 64.6 \\
  \rowcolor{mygray}
  Ours & InceptionV3 & \textbf{65.4} \\
  \bottomrule  
  \end{tabular}
  \vspace{-1.0em}
\end{table}
\addtolength{\tabcolsep}{-15pt} 

% \begin{figure*}[t]
%   \centering
%   \includegraphics[width=1.0\linewidth]{./images/loc_map_2.pdf}
%   % \vspace{-0.5em}
%   \caption{Qualitative object localization results compared with the baseline method. The predicted bounding boxes are in green, and the ground-truth boxes are in red.}
%   \label{loc_map}
%   % \vspace{-1.0em}
% \end{figure*}

\subsection{Ablation Study}
To demonstrate the effectiveness of counterfactual representation and decoupled loss, we conduct several ablation studies on the CUB-200-2011~\cite{wah2011caltech} as presented in Table~\ref{ablation_study_network}.

\textbf{Counterfactual Representation.} Training the baseline model with counterfactual representation results in significant improvements across all evaluation metrics. Specifically, when VGG16~\cite{simonyan2014very} serves as the backbone, counterfactual representation leads to an additional $3.6\%$, $5.2\%$, and $2.6\%$ improvement in Top-1 Cls, Top-1 Loc, and GT-known accuracy, respectively. For InceptionV3~\cite{szegedy2016rethinking} as the backbone, counterfactual representation achieves an extra $3.1\%$ improvement in Top-1 Cls and a $2.0\%$ improvement in Top-1 Loc compared to the baseline. Notably, when combining with Resnet50~\cite{he2016deep} as the backbone, it exhibits additional improvements of $4.1\%$, $6.1\%$, and $2.7\%$ in Top-1 Cls, Top-1 Loc, and GT-known accuracy, respectively.

\textbf{Decoupled Loss.} The application of the decoupled loss (Eq.\ref{eq:decouple}) respectively results in improvements of $0.2\%$, $0.4\%$, and $0.2\%$ in Top-1 Cls, Top-1 Loc, and GT-known accuracy when using VGG16~\cite{simonyan2014very} as the backbone. If the backbone is InceptionV3~\cite{szegedy2016rethinking}, training the model with the decoupled loss leads to additional improvements of $1.1\%$ and $1.2\%$ in Top-1 Loc and GT-known accuracy, respectively. If the backbone is Resnet50\cite{he2016deep}, it also brings additional improvements of $0.2\%$ and $0.1\%$ in Top-1 Cls and Top-1 Loc accuracy.

These ablation studies confirm the effectiveness of counterfactual representation and decoupled loss.

\addtolength{\tabcolsep}{-2pt} 
\begin{table}[t]
% \begin{wraptable}{r}{10cm}
   \centering
   \caption{Ablation studies on the CUB-200-2011. Base: NL-CCAM~\cite{yang2020combinational}, Count: counterfactual representation, Decou: decoupled loss.}
   \label{ablation_study_network}
  %  \vspace{0.5em}
  %  \scalebox{0.95}{
    \begin{tabular}{ccccccc}
      \toprule
     Backbone & Base & Count & Decou & Top-1 Cls  & Top-1 Loc  & GT-known \\ 
      \midrule
      % \hline
      \multirow{3}*{VGG16} & $\surd$ &  & & 75.6 & 68.5 & 90.0  \\
       & $\surd$  & $\surd$& & 79.2 & 73.7 & 92.6  \\
       & $\surd$ & $\surd$  & $\surd$  & \textbf{79.4} & \textbf{74.1}  & \textbf{92.8} \\
       \hline
       \multirow{3}*{InceptionV3} & $\surd$ & & & 79.1 & 75.2 & 94.6 \\
        &  $\surd$ & $\surd$& & \textbf{82.2} & 77.2 & 93.8 \\
        & $\surd$ & $\surd$  & $\surd$  & \textbf{82.2} & \textbf{78.3} & \textbf{95.0} \\
      \hline
      
      \multirow{3}*{Resnet50} &$\surd$ & & &80.6 &75.5 &93.3\\
      &$\surd$  &$\surd$ & &84.7 &81.6 &\textbf{96.0}  \\
      & $\surd$ & $\surd$  & $\surd$ & \textbf{84.9} & \textbf{81.7} & \textbf{96.0} \\
      % \hline
      \bottomrule
    \end{tabular}
    % \vspace{-1.0em}
\end{table}
\addtolength{\tabcolsep}{2pt} 

\subsection{Test-time Adaptation Experiment}
\addtolength{\tabcolsep}{-0.8pt}  
\begin{table}[t]
   \centering
   \caption{Adaptation comparison with different adaptation on the CUB-200-2011. Ada: adaptation.}
   \label{test-time analysis}
      \begin{tabular}{lcccc}
      \toprule
      \multirow{2}*{Method}      &
      \multicolumn{2}{c}{VGG16}&
      \multicolumn{2}{c}{InceptionV3} \\
      \cmidrule (r){2-3} \cmidrule (r){4-5}
      & Top-1 Loc & GT-known & Top-1 Loc & GT-known\\
      \midrule
      Without Ada           &  74.1  & 92.8  &  78.3 & 95.0 \\
      Ada with tent~\cite{wang2020tent}              &    74.4  & \textbf{94.0}       &  78.7   & 95.8 \\
      \rowcolor{mygray} Ada with ours          &    \textbf{74.9} & \textbf{94.0}  & \textbf{78.9} & \textbf{96.0}   \\
      \bottomrule
      \end{tabular}
      \vspace{-1.0em}
\end{table}
\addtolength{\tabcolsep}{0.8pt}

To highlight the importance of test-time adaptation, we conduct experiments on the CUB-200-2011 dataset~\cite{wah2011caltech}. In Table~\ref{test-time analysis}, we observe significant improvements in localization performance with both types of adaptation approaches. Furthermore, our adaptation approach outperforms tent~\cite{wang2020tent} comprehensively. Specifically, when using VGG16~\cite{simonyan2014very} as the backbone, our adaptation approach achieves an additional $0.8\%$ and $1.2\%$ improvement in Top-1 Loc and GT-known accuracy, respectively, outperforming tent by $0.5\%$ in Top-1 Loc accuracy. Similarly, when using InceptionV3~\cite{szegedy2016rethinking} as the backbone, our adaptation approach yields an additional $0.6\%$ and $1.0\%$ improvement in Top-1 Loc and GT-known accuracy, respectively, outperforming tent by $0.2\%$ in both metrics. These results underscore the superior performance of our adaptation approach compared to the tent in terms of improving localization accuracy when applied during test-time in the proposed Counterfactual-CAM.

\subsection{Robustness Analysis}
\addtolength{\tabcolsep}{-2.0pt} 
\begin{table}[t]
   \centering
   \caption{Performance using CAM~\cite{zhou2016learning} as our baseline on the
CUB-200-2011 dataset. AccV2: MaxBoxAccV2.}
   \label{tab:robustness}
   \begin{tabular}{lccccc}
   \toprule
   Method & Backbone & \makecell{Top-1 Cls} & \makecell{Top-1 Loc} & GT-known & AccV2\\
   \midrule
   CAM~\cite{zhou2016learning}  &VGG16    &  76.3        &    56.7  &  71.3 & 43.3 \\ 
   \rowcolor{mygray}CAM+Ours&VGG16 & \textbf{79.7} &	\textbf{68.4} &	\textbf{84.0} & \textbf{58.1} \\ 
   \hline
   CAM~\cite{zhou2016learning}   &InceptionV3    &   77.6        &    59.1  & 74.9 & 48.7 \\ 
   \rowcolor{mygray}CAM+Ours&InceptionV3 &\textbf{81.5} & \textbf{66.6}	& \textbf{81.2} & \textbf{52.3} \\
   \hline
  CAM~\cite{zhou2016learning}  &Resnet50    &  79.7        &    61.5  &  75.6 & 47.3 \\ 
   \rowcolor{mygray}CAM+Ours&Resnet50 &\textbf{84.8}&\textbf{71.1}&\textbf{82.2} & \textbf{55.5} \\	
   \hline

   \hline
   \end{tabular}
   \vspace{-1.0em}
\end{table}
\addtolength{\tabcolsep}{2.0pt} 

To further underscore the robustness and effectiveness of the Counterfactual Co-occurring Learning (CCL) in the WSOL task, we set up some additional experiments by substituting the original baseline model (\eg, NL-CCAM~\cite{yang2020combinational}) with CAM~\cite{zhou2016learning} and employing different backbone networks, as illustrated in Table~\ref{tab:robustness}. These results demonstrate that the Counterfactual Co-occurring Learning (CCL) consistently brings significant detection and segmentation performance improvements in the WSOL task.

\subsection{Computational Overhead Analysis}
\addtolength{\tabcolsep}{-1.0pt} 
\begin{table*}[t]
   \centering
   \caption{Computational overheads of our method during training and inference, respectively. Our experimental setup utilizes a batch size of $12$ and a single GeForce RTX $2080$ Ti GPU.}
   \label{tab:computational_overheads}
   \begin{tabular}{lccccccccccc}
   \toprule
   Method & \makecell{Madd\\(GMAdd)} & \makecell{Flops\\(GFlops)} & \makecell{MemR+W\\(MB)} & \makecell{Param\\(MB)} & \makecell{Training\\GPU Memory\\(MB)} & \makecell{Training\\Speed\\(img/s)} & \makecell{Inference\\GPU Memory\\(MB)} & \makecell{Inference\\Speed\\(img/s)} & \makecell{Top-1\\Cls} & \makecell{Top-1\\Loc} & \makecell{GT-\\known}\\
   \midrule
   VGG16-Baseline  &31.66    &  15.85        &   305.8  &   16.2 & 7131 &  61 & 3833 & 136 & 76.0 & 69.0 &  90.5\\ 
   \rowcolor{mygray}VGG16-Ours  &35.36    &  17.70        &   346.4  &   25.6 & 7261 &  53 & 3905 & 131 & \textbf{79.4} & \textbf{74.1} &  \textbf{92.8}\\ 
\hline
InceptionV3-Baseline  &10.89    &   5.45        &    220.5  &   10.7 & 3567 &  77 & 2063 &  129 & 79.1 & 75.2 &   94.6\\ 
\rowcolor{mygray}InceptionV3-Ours  &19.07    &  9.54        &   282.4  &   24.9 & 3745 &  60 & 2383 & 122 & \textbf{82.2} & \textbf{78.3} &  \textbf{95.0}\\ 
   \hline

   \hline
   \end{tabular}
   \vspace{-1.0em}
\end{table*}
\addtolength{\tabcolsep}{1.0pt} 

% We conduct a comparison experiment of speed and computational complexity in Table~\ref{tab:computational_overheads}. The results vividly depict that our method's GPU memory utilization for both training and inference remains quite similar to that of the baseline method. Similarly, the inference speed of our method closely mirrors that of the baseline method. Therefore, we believe that our method excels, showcasing substantial improvements in classification and localization while maintaining computational complexity similar to the baseline method.

We only collect features and pair them within each batch. The number of foreground features F, background features B, and counterfactual representations FB are batch size (bz), bz, and bz $\times$ bz, respectively. The training and computational costs are detailed in Table~\ref{tab:computational_overheads}, showing that using a small batch size does not notably increase the training time. Therefore, we believe that our method excels, showcasing substantial improvements in classification and localization while maintaining computational complexity similar to the baseline method.

\section{Conclusion}
In this paper, we undertake an early effort to address the ``biased activation'' issue arising from co-occurring backgrounds through the Counterfactual Co-occurring Learning (CCL) paradigm. Specifically, we introduce a counterfactual representation perturbation mechanism comprising co-occurring feature disentangling and counterfactual representation synthesizing. The former aims to separate the foreground and its co-occurring background from the original image. The latter involves synthesizing counterfactual representations by pairing the constant foreground with various backgrounds. By aligning predictions between the original image representations and counterfactual representations, we guide the detection model to concentrate on constant foreground information, disregarding diverse background information. Consequently, we remove the impact of co-occurring backgrounds, effectively addressing the ``biased localization'' problem.

\section*{Acknowledgement}
% This work was supported by the National Natural Science Foundation of China (62337001), National Natural Science Foundation of China (62293554, 62206249, U2336212), Natural Science Foundation of Zhejiang Province, China (LZ24F020002), Young Elite Scientists Sponsorship Program by CAST (2023QNRC001), and the Fundamental Research Funds for the Central Universities(No. 226-2022-00051).

This work was supported by the National Natural Science Foundation of China (62337001, 62293554, 62206249, U2336212), Natural Science Foundation of Zhejiang Province, China (LZ24F020002), Young Elite Scientists Sponsorship Program by CAST (2023QNRC001), and the Fundamental Research Funds for the Central Universities(No. 226-2022-00051).

\bibliographystyle{IEEEtran}
\bibliography{IEEEabrv,main}

\end{document}